\documentclass{article}

\usepackage{hyperref}

\usepackage[utf8]{inputenc} 
\usepackage[T1]{fontenc}    
\usepackage{url}            
\usepackage{booktabs}       
\usepackage{amsfonts}       
\usepackage{nicefrac}       
\usepackage{microtype}      
\usepackage{xcolor}         

\usepackage{graphicx}
\usepackage{amsmath}
\usepackage{amssymb}
\usepackage{booktabs}
\usepackage[utf8]{inputenc} 
\usepackage[T1]{fontenc}    
\usepackage{url}            
\usepackage{amsfonts}       
\usepackage{nicefrac}       
\usepackage{microtype}      
\usepackage{bbding}
\usepackage{pifont}
\usepackage{bbm}
\usepackage{soul}
\usepackage{multirow}
\usepackage{framed}
\usepackage{amsthm}
\usepackage{makecell}

\usepackage{amsmath}
\usepackage{amssymb}
\usepackage{mathtools}
\usepackage{amsthm}
\allowdisplaybreaks[4]

\usepackage{wrapfig}
\usepackage{caption}

\usepackage{amsmath}
\usepackage{amssymb}
\usepackage{mathtools}
\usepackage{amsthm}

\usepackage[capitalize,noabbrev]{cleveref}

\theoremstyle{plain}

\theoremstyle{definition}

\theoremstyle{remark}

\usepackage[textsize=tiny]{todonotes}

\usepackage[accepted]{icml2023}

\icmltitlerunning{Does a Neural Network Really Encode Symbolic Concepts?}

\begin{document}

\twocolumn[
\icmltitle{Does a Neural Network Really Encode Symbolic Concepts?}




\begin{icmlauthorlist}
\icmlauthor{Mingjie Li}{sjtu}
\icmlauthor{Quanshi Zhang}{sjtu}
\end{icmlauthorlist}

\icmlaffiliation{sjtu}{Shanghai Jiao Tong University}

\icmlcorrespondingauthor{Quanshi Zhang. Quanshi Zhang is the corresponding author. He is with the Department of Computer Science and Engineering, the John Hopcroft Center, at the Shanghai Jiao Tong University, China.}{zqs1022@sjtu.edu.cn}

\icmlkeywords{Machine Learning, ICML}

\vskip 0.3in
]



\printAffiliationsAndNotice{\icmlEqualContribution} 

\begin{abstract}
Recently, a series of studies have tried to extract interactions between input variables modeled by a DNN and define such interactions as concepts encoded by the DNN.
However, strictly speaking, there still lacks a solid guarantee whether such interactions indeed represent meaningful concepts.
Therefore, in this paper, we examine the trustworthiness of interaction concepts from four perspectives.
Extensive empirical studies have verified that a well-trained DNN usually encodes sparse, transferable, and discriminative concepts, which is partially aligned with human intuition.
The code is released at \url{https://github.com/sjtu-xai-lab/interaction-concept}.
\end{abstract}

\section{Introduction}

Understanding the black-box representation of deep neural networks (DNNs) has received increasing attention in recent years.
Unlike graphical models with interpretable internal logic, the layerwise feature processing in DNNs makes it naturally difficult to explain DNNs from the perspective of symbolic concepts.
Instead, previous studies interpreted DNNs from other perspectives, such as illustrating the visual appearance that maximizes the inference score~\cite{simonyan2013deep,yosinski2015understanding}, and estimating attribution/importance/saliency of input variables~\cite{ribeiro2016should,sundararajan2017axiomatic,lunberg2017unified}.
\citet{zhou2014object,bau2017network,kim2018interpretability} visualized the potential correspondence between convolutional filters in a DNN and visual concepts in an empirical manner.

\textbf{Unlike previous studies, a series of studies~\cite{ren2021towards,ren2021learning,deng2022discovering} tried to define and propose an exact formulation for the concepts encoded by a DNN.}
Specifically, these studies discovered that a well-trained DNN usually encoded various interactions between different input variables, and the inference score on a specific input sample could be explained by numerical effects of different interactions.
Thus, they claimed that each interaction pattern was a symbolic concept encoded by the DNN.

Specifically, let us consider a DNN given a sample with {\small$n$} input variables {\small$N=\{1,2,...,n\}$} \emph{e.g.}, given a sentence with five words ``\textit{he is a green hand.}''
The DNN usually does not use each individual input variable for inference independently.
Instead, the DNN lets different input variables interact with each other to construct concepts for inference.
For example, a DNN may memorize the interaction between words in {\small$S=\{\textit{green, hand}\}$} with a specific numerical contribution {\small$I(S)$} to push the DNN's inference towards the meaning of a ``\textit{beginner}.''
Such a combination of words is termed an \textit{interaction concept}.
Each interaction concept {\small$S\subseteq N$} represents the AND relationship between input variables in {\small$S$}.
Only the co-appearance of input variables in {\small$S$} can make an interaction effect {\small$I(S)$} on the network output.
In contrast, masking any word in {\small$\{\textit{green, hand}\}$} removes the interaction effect {\small$I(S)$}.
In this way, \citet{ren2021towards} proved that the inference score {\small$y$} of a trained DNN on each sample can be written as the sum of effects of all potential symbolic concepts {\small$S\subseteq N$}, \emph{i.e.} {\small$y=\sum_{S\subseteq N}I(S)$}.

\begin{figure*}[t]
    \centering
    \includegraphics[width=.95\linewidth]{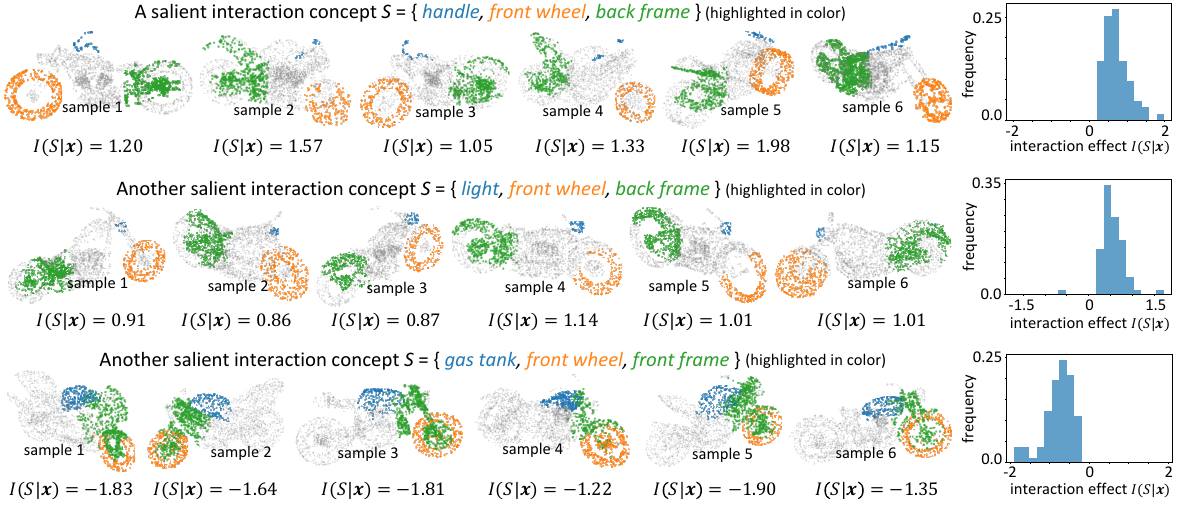}
    \vspace{-8pt}
    \caption{Visualization of interaction concepts {\scriptsize$S$} extracted by PointNet on different samples in the ShapeNet dataset.
    The histograms show the distribution of interaction effects {\scriptsize$I(S|\boldsymbol{x})$} over samples in the ``motorbike'' category, where {\scriptsize$S$} is extracted as a salient concept.}
    \label{fig:visualize-concepts-pointcloud}
    \vspace{-5pt}
\end{figure*}

\textit{\textbf{However, the claim that ``a DNN encodes symbolic concepts'' is too counter-intuitive.
\citet{ren2021towards} just formulated {\small$I(S)$} that satisfied {\small$y=\sum_{S\subseteq N}I(S)$}. 
Current studies have not provided sufficient support for the claim that a DNN really learns symbolic concepts.}
In fact, we should not ignore another potential situation that the defined effect {\small$I(S)$} is just a mathematical transformation that ensures the decomposition of network output into concepts {\small$y=\sum_{S\subseteq N}I(S)$}, rather than faithfully representing a meaningful and transferable concept learned by a DNN.}

Therefore, in this paper, we aim to give a quantitative verification of the concept-emerging phenomenon, \emph{i.e.}, whether a well-trained DNN summarizes transferable symbolic knowledge from chaotic raw data, like what human brain does.
Or the defined effect {\small$I(S)$} is just a mathematical game without clear meanings.
To this end, we believe that if a well-trained DNN really encodes certain concepts, then the concepts are supposed to satisfy the following four requirements.

$\bullet$~\textbf{Representing network inference using a few concepts.} 
If a DNN really learns symbolic concepts, then the DNN's inference on a specific sample is supposed to be concisely explained by a small number of salient concepts, rather than a large number of concepts, according to both Occam’s Razor and people's intuitive understanding towards concepts.
In fact, the sparsity of concepts in a specific sample has been discussed by \citet{ren2021towards}. 
In this paper, we further conduct extensive experiments on more diverse DNNs, in order to verify that a well-trained DNN usually extracts sparse concepts from each sample for inference.

$\bullet$~\textbf{A transferable concept dictionary through different samples.}
If a DNN can use a relatively small set {\small$\mathbf{D}$} of salient concepts, namely a concept dictionary, to approximate inference scores on different samples in a category, \emph{i.e.}, {\small$\forall\boldsymbol{x},y\approx \sum_{S\in\mathbf{D}}I(S|\boldsymbol{x})$}, then we consider the concept dictionary {\small$\mathbf{D}$} represents common features shared by different samples in the category.
Otherwise, if \citet{ren2021towards} extract a fully different set of concepts from each different sample in the same category, then these concepts probably represent noisy signals.
In other words, convincing concepts must be stably extracted with high transferability through different samples in the same category.

$\bullet$~\textbf{Transferability of concepts across different DNNs.}
Similarly, when we train different DNNs for the same task, different DNNs probably learn  similar sets of concepts if they really memorize the defined ``concepts'' as basic inference patterns for the task.

$\bullet$~\textbf{Discrimination power of concepts.}
Furthermore, if a DNN learns meaningful concepts, then these concepts are supposed to exhibit certain discrimination power in the classification task.
The same concept extracted from different samples needs to consistently push the DNN towards the classification of a certain category.

To this end, we conducted experiments on various DNNs trained on different datasets for classification tasks, including tabular datasets, image datasets, and point-cloud datasets.
We found that all these trained DNNs encoded transferable concepts.
On the other hand, we also investigated a few extreme cases, in which the DNN either collapsed into simple linear models or failed to learn transferable and discriminative concepts.
In sum, although we cannot theoretically prove the phenomenon of the emergence of transferable concepts, this phenomenon indeed happened for most tasks in our experiments.

\textbf{Contributions} of this paper can be summarized as follows.
(1) Besides the sparsity of concepts, we propose three more perspectives to examine the concept-emerging phenomenon of a DNN, \emph{i.e.}, whether the DNN summarizes transferable symbolic knowledge from chaotic raw data.
(2) Extensive empirical studies on various tasks have verified that a well-trained DNN usually encodes transferable interaction concepts.
(3) We also discussed three extreme cases, in which a DNN is unlikely to learn transferable concepts.

\vspace{-2pt}
\section{Related works}
\vspace{-2pt}

\subsection{Understanding black-box representation of DNNs}
Many explanation methods have been proposed to explain DNNs from different perspectives.
Typical explanation methods include visualizing patterns encoded by a DNN~\cite{simonyan2013deep,zeiler2014visualizing,yosinski2015understanding,dosovitskiy2016inverting}, estimating the attribution/importance/saliency of each input variable~\cite{ribeiro2016should,sundararajan2017axiomatic,lundberg2017unified,fong2017interpretable,zhou2016learning,selvaraju2017grad}, and learning feature vectors potentially correspond to semantic concepts~\cite{kim2018interpretability}.
Unlike attribution methods, some studies focused on quantifying interactions between input variables~\cite{sorokina2008detecting,murdoch2018beyond,singh2018hierarchical,jin2019towards,janizek2020explaining}.
In game theory, \citet{grabisch1999axiomatic,sundararajan2020shapley,tsai2022faith} proposed interaction metrics from different perspectives.
Some studies explained a DNN by distilling the DNN into another interpretable model~\cite{frosst2017distilling,che2016interpretable,wu2018beyond,zhang2018interpreting,vaughan2018explainable,tan2018learning}.
However, most explanation methods did not try to disentangle concepts encoded by a DNN.

\subsection{Explainable AI (XAI) theories based on game-theoretic interactions}

Our research group developed a theoretical framework based on game-theoretic interactions, which aims to tackle the following two challenges in XAI, \emph{i.e.}, (1) extracting and quantifying concepts from implicit knowledge representations of DNNs and (2) utilizing these explicit concepts to explain the representational capacity of DNNs.
Furthermore, we discovered that game-theoretic interactions provide a new perspective for analyzing the common underlying mechanism shared by previous XAI applications.

$\bullet$ \textit{Using game-theoretical interactions to define concepts encoded by DNNs.}
Quantifying the interactions between input variables is one of the ultimate problems facing XAI~\cite{sundararajan2020shapley,tsai2022faith}.
Based on game theory, we introduced multi-variate interactions \cite{zhangdie2021building, zhanghao2021interpreting} and multi-order interactions \cite{zhang2020interpreting} to analyze interactions encoded by the DNN.
Recently, \citet{ren2021towards} proposed the mathematical formulation for concepts encoded by a DNN, and \citet{ren2021learning} further used such concepts to define the optimal baseline values for Shapely values.
Based on this, recent studies have also observed~\cite{ren2021learning} and mathematically proved~\cite{ren2023we} the concept-emerging phenomenon in DNNs.
\textbf{However, strictly speaking, there still lacks theoretical guarantee for the interaction to prove whether the interaction represents the true concept encoded by a DNN or just a tricky metric without a clear meaning.}
\textbf{Therefore, in this study, we examined the trustworthiness of the interaction concepts from four perspectives.}

$\bullet$ \textit{Explaining the representation power of DNNs based on game-theoretic interactions.}
Game-theoretical interactions facilitate the explanation of the representation capacity of a DNN from different perspectives, including the adversarial robustness~\cite{wangxin2021interpreting,ren2021adversarial}, adversarial transferability~\cite{wangxin2021unified}, and generalization power~\cite{zhang2020interpreting, zhouhuilin2023generalization}.
Besides, the game-theoretical interactions can also be utilized to explain the signal processing behavior of DNNs.
For example, \citet{chengxu2021concepts} analyzed the distinctive behavior of a DNN encoding shape/texture features based on such interactions.
\citet{chengxu2021hypothesis} discovered that salient interactions often represented different prototype features encoded by a DNN.
\citet{deng2022discovering} proved that it was difficult for a DNN to encode mid-order interactions, which reflected a representation bottleneck of DNNs.
In comparison, \citet{deng2023BNN} proved that a Bayesian neural network was less likely to encode high-order interactions, thereby alleviating the over-fitting problem.

$\bullet$ \textit{Unifying empirical findings in the framework of game-theoretic interactions.}
To unify different attribution methods, \citet{deng2022unify} used interactions as a unified reformulation of different attribution methods. They proved that attributions estimated by each of 14 attribution methods could all be represented as a certain allocation of interaction effects to different input variables.
In addition, \citet{zhangquanshi2022proving} proved that the reduction of interactions was the common mathematical mechanism shared by a total of 12 previous approaches to enhance adversarial transferability.

\begin{figure*}[t]
    \centering
    \includegraphics[width=.95\linewidth]{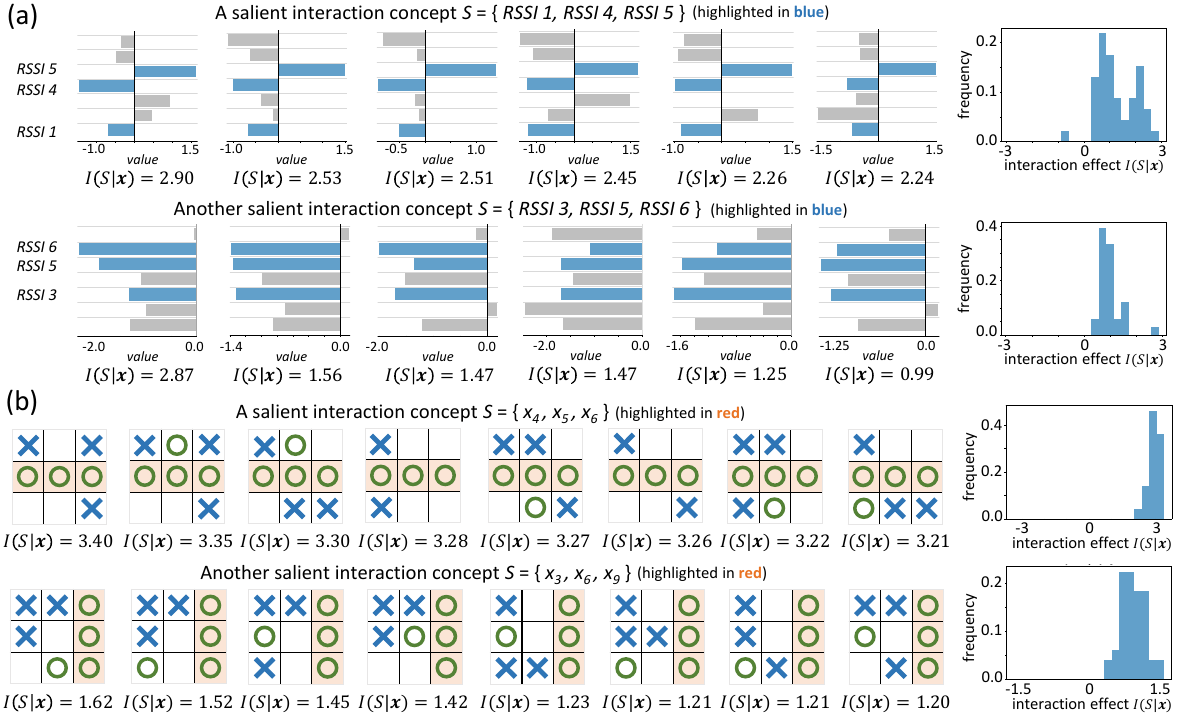}
    \vspace{-10pt}
    \caption{Visualization of interaction concepts {\scriptsize$S$} extracted by two \textit{MLP-5} networks\footref{fn:exp-setting}, which are trained on (a) the \textit{wifi} dataset\footref{fn:exp-setting} and (b) the \textit{tic-tac-toe} dataset\footref{fn:exp-setting}.
    The histograms show (a) the distribution of interaction effects {\scriptsize$I(S|\boldsymbol{x})$} over samples in the $4^{\text{th}}$ category, and (b) the distribution of interaction effects {\scriptsize$I(S|\boldsymbol{x})$} over samples in sub-categories\footref{fn:specific-category} with patterns {\scriptsize$x_4\!=\!x_5\!=\!x_6\!=\!1$} and {\scriptsize$x_3\!=\!x_6\!=\!x_9\!=\!1$}.}
    \label{fig:visualize-concepts-tabular}
    \vspace{-10pt}
\end{figure*}

\section{Emergence of transferable concepts}
\label{sec:emergence}

\subsection{Preliminaries: representing network inferences using interaction concepts}
\label{subsec:preliminaries}

It is widely believed that the learning of a DNN can be considered as a regression problem, instead of explicitly learning symbolic concepts like how graphical models do.
However, a series of studies~\cite{ren2021towards,ren2021learning,deng2022discovering} have discovered that given a sufficiently-trained DNN for a classification task, its inference logic on a certain sample can usually be rewritten as the detection of specific concepts.
In other words, the DNN's inference score on a specific sample can be sparsely disentangled into effects of a few concepts.

\textbf{Disentangling the DNN's output as effects of interaction concepts.}
Specifically, let us consider a trained DNN {\small$v:\mathbb{R}^n\to\mathbb{R}$} and an input sample {\small$\boldsymbol{x}$} with {\small$n$} input variables indexed by {\small$N=\{1,2,...,n\}$}.
Here, we assume the network output {\small$v(\boldsymbol{x})\in\mathbb{R}$} is a scalar.
Note that different settings can be applied to {\small$v(\boldsymbol{x})$}.
For example, for multi-category classification tasks, we may set {\small$v(\boldsymbol{x})=\log\frac{p(y=y^{\text{truth}}|\boldsymbol{x})}{1-p(y=y^{\text{truth}}|\boldsymbol{x})}\in\mathbb{R}$} by following~\cite{deng2022discovering}.
Then, given a function {\small$v(\boldsymbol{x})$}, \citet{ren2021towards} have proposed the following metric to quantify the interaction concept that is comprised of input variables in {\small$S\subseteq N$}.

\vspace{-7pt}
\begin{small}
\begin{equation}
    I(S|\boldsymbol{x}) \triangleq {\sum}_{T\subseteq S} (-1)^{|S|-|T|}\cdot v(\boldsymbol{x}_T)
    \label{eq:harsanyi-def}
\end{equation}
\end{small}

\vspace{-7pt}
where {\small$\boldsymbol{x}_T$} denotes the input sample when we keep variables in {\small$T\subseteq N$} unchanged and mask variables in {\small$N\backslash T$} using baseline values\footnote{For all tabular datasets and the image datasets (the \textit{CelebA-eyeglasses} dataset and the \textit{CUB-binary} dataset), the baseline value of each input variable was set as the mean value of this variable over all samples~\cite{dabkowski2017real}. For grayscale digital images in the \textit{MNIST-3} dataset, the baseline value of each pixel was set as zero~\cite{ancona2019explaining}. For the point-cloud dataset, the baseline value was set as the center of the entire point cloud~\cite{shen2021interpreting}.}.

\textbf{Here, the interaction concept {\small$I(S|\boldsymbol{x})$} extracted from the input {\small$\boldsymbol{x}$} encodes an AND relationship (interaction) between input variables in {\small$S$}.}
For example, let us consider three image regions of {\small$S=\{\textit{eyes, nose, mouth}\}$} that form the \textit{``face''} concept in a face classification task. 
Then, {\small$I(S|\boldsymbol{x})$} measures the numerical effect of the concept on the classification score {\small$v(\boldsymbol{x})$}.
Only when all image regions of \textit{``eyes''}, \textit{``nose''}, and \textit{``mouth''} co-appear in the input image, the \textit{``face''} concept is activated, and contributes a numerical effect {\small$I(S|\boldsymbol{x})$} to the classification score.
Otherwise, if any region is masked, then the \textit{``face''} concept cannot be formed, which removes the interaction effect, making {\small$I(S|\boldsymbol{x}_{\text{masked}})=0$}.

Mathematically, the above definition for an interaction concept can be understood as the Harsanyi dividend~\cite{harsanyi1963simplified} of {\small$S$} \emph{w.r.t.} the DNN {\small$v$}.
It has been proven that the DNN's inference score {\small$v(\boldsymbol{x})$} can be disentangled into the sum of effects of all potential concepts, as follows.

\vspace{-8pt}
\begin{small}
\begin{equation}
    v(\boldsymbol{x})={\sum}_{S\subseteq N} I(S|\boldsymbol{x})
    ~\approx{\sum}_{S\in\Omega_{\boldsymbol{x}}} I(S|\boldsymbol{x})
    \label{eq:harsanyi-sum}
\end{equation}
\end{small}

\vspace{-7pt}
In particular, all interaction concepts can be further categorized into a set of \textbf{salient concepts} {\small$S\in\Omega_{\boldsymbol{x}}$} with considerable effects {\small$I(S|\boldsymbol{x})$}, and a set of ignorable \textbf{noisy patterns} with almost zero effects {\small$I(S|\boldsymbol{x})\approx 0$}.

Note that the Harsanyi dividend {\small$I(S|\boldsymbol{x})$} also satisfies many desirable axioms/theorems, as introduced in Appendix \ref{app:harsanyi-axiom-theorem}.
The interaction effects can be further optimized using the trick of disentangling OR interactions\footref{fn:or-trick} introduced in both~\cite{li2023defining} and Appendix H.4 of \cite{ren2021towards}, to pursue higher sparsity of interaction concepts.

\begin{figure*}[t]
    \centering
    \includegraphics[width=.95\linewidth]{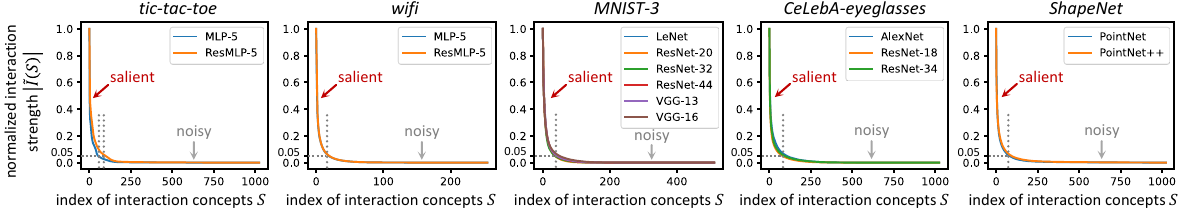}
    \vspace{-10pt}
    \caption{Normalized strength of interaction effects of different concepts in a descending order.
    DNNs trained for different tasks all encode sparse salient concepts.}
    \label{fig:sparsity-single-sample}
    \vspace{-8pt}
\end{figure*}

\subsection{Visualization of interaction concepts}
\label{subsec:visualize-concept}

In this section, we visualize interaction concepts extracted from point-cloud data and tabular data.
Note that a sample in the ShapeNet dataset~\cite{yi2016scalable} usually contains 2500 3D points.
To simplify the visualization, we simply consider 8-10 semantic parts on the point cloud {\small$\boldsymbol{x}$}, which has been provided by the dataset.\footnote{For example, the ShapeNet dataset has provided the annotated parts for the \textit{motorbike} category, including \textit{gas tank}, \textit{seat}, \textit{handle}, \textit{light}, \textit{front wheel}, \textit{back wheel}, \textit{front frame}, \textit{mid frame}, and \textit{back frame}. Please see Appendix~\ref{app:annotation} for details on the annotation of semantic parts.}
Each semantic part is taken as a ``single'' input variable to the DNN.
In this way, we visualize concepts consisting of semantic parts.

Fig.~\ref{fig:visualize-concepts-pointcloud} shows interaction concepts {\small$S$} and the corresponding effects {\small$I(S|\boldsymbol{x})$} extracted by PointNet~\cite{qi2017pointnet} from different samples {\small$\boldsymbol{x}$} in the ShapeNet dataset.
We find that the interaction concept {\small$S=\{\textit{light, front wheel, mid frame}\}$} on five samples all makes positive effects {\small$I(S|\boldsymbol{x})>0$} to the PointNet's output, whereas the interaction concept {\small$S=\{\textit{handle, front wheel, front frame}\}$} usually makes negative effects {\small$I(S|\boldsymbol{x})<0$} to the PointNet's output.
Similarly, Fig.~\ref{fig:visualize-concepts-tabular} shows interaction concepts extracted from two tabular datasets, \emph{i.e.}, the \textit{wifi} dataset\footref{fn:exp-setting}, and the \textit{tic-tac-toe} dataset\footref{fn:exp-setting}.
We visualize interactions between different \textit{received signal strength indicatons} (RSSIs) in the \textit{wifi} dataset, and interactions between \textit{board states} in the \textit{tic-tac-toe} dataset.
We also find that the same interaction concept usually makes similar effects to the network output on different input samples, which supports the conclusion in Section \ref{subsubsec:trans-different-sample}.

\subsection{Does a DNN really learn symbolic concepts?}

Although~\cite{ren2021towards} have claimed that the metric {\small$I(S|\boldsymbol{x})$} in Eq. (\ref{eq:harsanyi-def}) quantifies symbolic concepts encoded by a DNN, there is still no theory to guarantee a subset {\small$S\subseteq N$} with a salient effect {\small$I(S|\boldsymbol{x})$} faithfully represents a meaningful and transferable concept.
Instead, we should not ignore the possibility that {\small$I(S|\boldsymbol{x})$} is just a mathematical transformation that ensures {\small$v(\boldsymbol{x})=\sum_{S\subseteq N}I(S|\boldsymbol{x})$} in Eq. (\ref{eq:harsanyi-sum}) on each specific sample.
Therefore, in this study, we examine the counter-intuitive conjecture  that a DNN learns symbolic concepts from the following four perspectives.

\begin{figure*}[t]
    \centering
    \vspace{-2pt}
    \includegraphics[width=.95\linewidth]{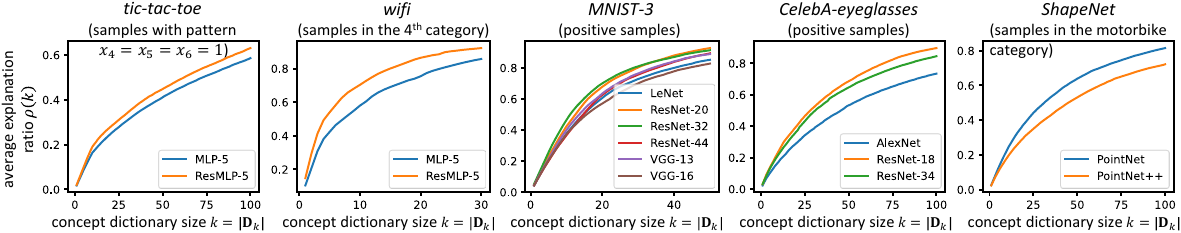}
    \vspace{-10pt}
    \caption{The change of the average explanation ratio {\small$\rho(k)$} along with the size {\small$k$} of the concept dictionary {\small$\mathbf{D}_k$}.}
    \label{fig:verify-concept-dictionary}
    \vspace{-10pt}
\end{figure*}

\subsubsection{Sparsity of the encoded concepts}
\label{subsubsec:sparsity-single-sample}

According to Eq. (\ref{eq:harsanyi-sum}), the DNN may encode at most {\small$2^n$} symbolic concepts in {\small$2^N\triangleq\{S:S\subseteq N\}$} \emph{w.r.t.} the {\small$2^n$} different combinations of input variables.
\textbf{However, a distinctive property of symbolism, which is different from connectionism, is that people usually would like to use \textit{a small number of explicit} symbolic concepts to represent the knowledge, instead of using \textit{extensive fuzzy} features.}

Thus, we hope to examine whether a DNN's inference score {\small$v(\boldsymbol{x})$} on a specific sample can be summarized into effects of a small number of salient concepts {\small$v(\boldsymbol{x})\approx \sum_{S\in\Omega_{\boldsymbol{x}}} I(S|\boldsymbol{x})$}, rather than using an exponential number of concepts \emph{w.r.t.} all subsets {\small$S\subseteq N$}.
To be precise, a faithful conceptual representation requires most concepts {\small$S\subseteq N$} to be noisy patterns with negligible effects {\small$I(S|\boldsymbol{x})\approx 0$}.
Only a few salient concepts {\small$S$} in {\small$\Omega_{\boldsymbol{x}}$} make considerable effects {\small$I(S|\boldsymbol{x})$}.

To this end, \citet{ren2021towards} have made a preliminary attempt to explain a DNN's inference score {\small$v(\boldsymbol{x})$} on a specific sample 
{\small$\boldsymbol{x}$} as interaction effects {\small$I(S|\boldsymbol{x})$} of a small number of concepts.
Specifically, they used a few top-ranked salient interaction concepts to explain the inference score of LSTMs~\cite{hochreiter1997long} and CNNs~\cite{kim-2014-convolutional} trained for sentiment classification and linguistic acceptance classification tasks on the SST-2 dataset~\cite{socher2013recursive} and the CoLA dataset~\cite{warstadt2019neural}.

\textbf{Experiments.}
In this paper, we further examined whether most DNNs, which were trained for much more diverse tasks on different datasets, all encoded very sparse salient concepts.
To this end, we trained various DNNs\footnote{Please see the \textit{experimental settings} paragraph at the end of Section~\ref{sec:emergence} for details on datasets and DNNs.\label{fn:exp-setting}} on tabular datasets (the \textit{tic-tac-toe} dataset\footref{fn:exp-setting} and the \textit{wifi} dataset\footref{fn:exp-setting}), image datasets (the \textit{MNIST-3} dataset\footref{fn:exp-setting} and the \textit{CelebA-eyeglasses} dataset\footref{fn:exp-setting}), and a point-cloud dataset (the \textit{ShapeNet} dataset\footref{fn:exp-setting}).
The interaction effects can be further optimized using the trick of disentangling OR interactions\footnote{This study also extracts the OR interaction, which is proved to be a specific AND interaction.\label{fn:or-trick}} introduced in both~\cite{li2023defining} and Appendix H.4 of \cite{ren2021towards}, to pursue higher sparsity of interaction concepts.
Fig.~\ref{fig:sparsity-single-sample} shows the normalized interaction strength of different concepts {\small$|\tilde{I}(S|\boldsymbol{x})|\!\triangleq\! |I(S|\boldsymbol{x})|/\max_{S'}|I(S'|\boldsymbol{x})|$} in a descending order for each DNN.
Each curve shows the strength averaged over different samples in the dataset.
We found that most concepts had little effects on the output {\small $|I(S|\boldsymbol{x})|\approx 0$}, which verified the sparsity of the encoded concepts.

\textbf{Salient concepts.} 
According to the above experiments, we can define the set of salient concepts as {\small$\Omega_{\boldsymbol{x}}=\{S:|I(S|\boldsymbol{x})|>\tau\}$}, subject to {\small$\tau=0.05\cdot\max_{S}|I(S|\boldsymbol{x})|$}.
As Fig.~\ref{fig:sparsity-single-sample} shows, there were only about 20-80 salient concepts extracted from an input sample, and all other concepts have ignorable effects on the network output.

\subsubsection{Transferability over different samples}
\label{subsubsec:trans-different-sample}

Beyond sparsity, the transferability of concepts is more important. 
If {\small$I(S|\boldsymbol{x})$} is just a tricky mathematical transformation on {\small$v(\boldsymbol{x}_S)$} without representing meaningful concepts, then each salient concept {\small$I(S|\boldsymbol{x})$} extracted from the input sample {\small$\boldsymbol{x}$} probably cannot be transferred to another input sample, \emph{i.e.,} we cannot extract the same salient concept consisting of variables in {\small$S$} in the second sample, due to sparsity of salient concepts.

Therefore, in this section, \textbf{we aim to verify whether concepts extracted from a sample can be transferred to other samples.} 
This task is actually equivalent to checking whether there exists a common concept dictionary, which contains most salient concepts extracted from different samples in the same category.

Given a well-trained DNN, we construct a relatively small dictionary {\small$\mathbf{D}_k\subseteq 2^N$} containing the top-{\small$k$} frequent concepts in different samples, and check whether such a dictionary contains most salient concepts in {\small$\Omega_{\boldsymbol{x}}$} extracted from each sample {\small$\boldsymbol{x}$}.
The concept dictionary {\small$\mathbf{D}_k$} is constructed based on a greedy strategy.
Specifically, we first extract a set of salient concepts {\small$\Omega_{\boldsymbol{x}}$} from each input sample {\small$\boldsymbol{x}$}.
Then, we compute the frequency of each concept {\small$S$} being a salient concept over different samples.
Finally, the concept dictionary {\small$\mathbf{D}_k$} is constructed to contain the top-{\small$k$} interaction concepts with the highest frequencies.

Then, we design the metric {\small$\rho(k)\triangleq\mathbb{E}_{\boldsymbol{x}}[|\mathbf{D}_k\cap\Omega_{\boldsymbol{x}}|/|\Omega_{\boldsymbol{x}}|]$} to evaluate the average ratio of concepts extracted from an input sample that is covered by the concept dictionary {\small$\mathbf{D}_k$}. 
Theoretically, if we construct a larger dictionary {\small$\mathbf{D}_k$} with more concepts (a larger {\small$k$} value), then the dictionary can explain more concepts.

\textbf{Experiments.} 
We conducted experiments to show whether there existed a small concept dictionary that could explain most concepts encoded by the DNN.
Specifically, we constructed a concept dictionary to explain samples in a certain category in each dataset\footnote{In this paper, when we needed to analyze of samples in a specific category, we used positive samples in the \textit{MNIST-3}, \textit{CelebA-eyeglasses}, and \textit{CUB-binary} datasets, samples in the {\small$4^{\text{th}}$} category in the \textit{wifi dataset}, and samples in the ``motorbike'' category in the ShapeNet dataset. For the \textit{tic-tac-toe} dataset, since there exists eight sub-categories among positive samples, we used samples in the sub-category with the pattern {\scriptsize$x_4=x_5=x_6=1$}.\label{fn:specific-category}}.
In this experiment, we temporarily extracted salient concepts using {\small$\tau=0.1\cdot \max_{S}|I(S|\boldsymbol{x})|$} to construct {\small$\Omega_{\boldsymbol{x}}$}
\footnote{Here, we increased the threshold from {\scriptsize$\tau\!=\!0.05\cdot \max_{S}|I(S|\boldsymbol{x})|$} to {\scriptsize$\tau\!=\!0.1\cdot \max_{S}|I(S|\boldsymbol{x})|$} to analyze those highly salient concepts. Appendix~\ref{app:concept-dict-more} shows results computed by using the vanilla threshold {\scriptsize$\tau\!=\!0.05\cdot \max_{S}|I(S|\boldsymbol{x})|$} to compute {\scriptsize$\Omega_{\boldsymbol{x}}$}.}.
Fig.~\ref{fig:verify-concept-dictionary} shows the increase of the average explanation ratio {\small$\rho(k)$} along with the increasing size {\small$k$} of the concept dictionary {\small$\mathbf{D}_k$}.
We found that there usually existed a concept dictionary consisting of 30-100 concepts, which could explain more than 60\%-80\% salient concepts encoded by the DNN.
Besides, Fig.~\ref{fig:visualize-concepts-pointcloud}(right) and Fig.~\ref{fig:visualize-concepts-tabular}(right) also show histograms of effects {\small$I(S|\boldsymbol{x})$} over different samples\footref{fn:specific-category}, where {\small$S$} was extracted as a salient concept.
We found that the same interaction concept usually made similar effects on different samples.
This verified that the DNN learned transferable concepts over different samples.

\begin{figure*}[t]
    \centering
    \includegraphics[width=.95\linewidth]{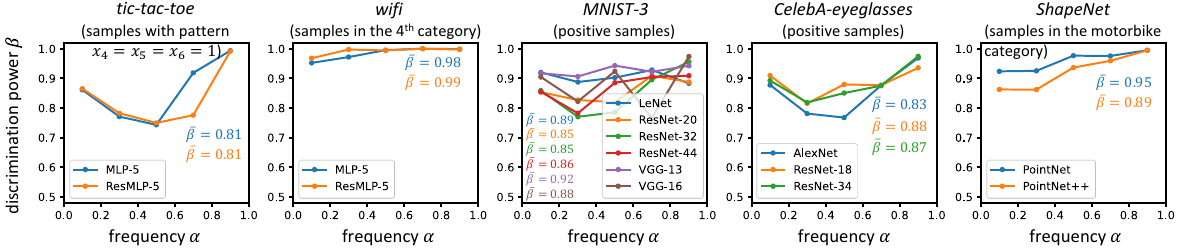}
    \vspace{-12pt}
    \caption{The average discrimination power of concepts in different frequency intervals, \emph{i.e.} {\small$\alpha\in(0.0,0.2],(0.2,0.4],...,(0.8,1.0]$}.
    The weighted average discrimination power {\small$\bar{\beta}$} over concepts of all frequencies is shown beside the curve.}
    \label{fig:concept-discrimination-power}
    \vspace{-10pt}
\end{figure*}

\subsubsection{Transferability across different DNNs}
\label{subsubsec:trans-different-dnn}

In addition to sample-wise transferability of concepts, another aspect is model-wise transferability.
If the concepts extracted from an input sample really represent meaningful knowledge for the task, then these concepts are supposed to be stably learned by different DNNs towards the same task, although we cannot directly align intermediate-layer features between different DNNs.

Therefore, in this section, \textbf{we aim to verify whether concepts extracted from a DNN can be transferred to another DNN trained for the same task.}
In other words, we actually check whether salient concepts encoded by one DNN are also encoded by another DNN learned for the same task.
To this end, let us consider two DNNs, {\small$v_1$} and {\small$v_2$}, trained for the same task.
Given an input sample {\small$\boldsymbol{x}$}, let {\small$\Omega^{v_1}_{\boldsymbol{x}}$} and {\small$\Omega^{v_2}_{\boldsymbol{x}}$} denote the sets of salient concepts extracted by {\small$v_1$} and {\small$v_2$} from the input sample {\small$\boldsymbol{x}$}, respectively.
We evaluate the the ratio of concepts in {\small$\Omega^{v_1}_{\boldsymbol{x}}$} encoded by {\small$v_1$}, which are also encoded by {\small$v_2$} in {\small$\Omega^{v_2}_{\boldsymbol{x}}$}, \emph{i.e.} {\small$\gamma(\Omega^{v_1}_{\boldsymbol{x}}|\Omega^{v_2}_{\boldsymbol{x}})\triangleq |\Omega^{v_1}_{\boldsymbol{x}}\cap \Omega^{v_2}_{\boldsymbol{x}}|/|\Omega^{v_1}_{\boldsymbol{x}}|$}, to measure the transferability of salient concepts in {\small$\Omega^{v_1}_{\boldsymbol{x}}$}.
A larger ratio {\small$\gamma(\Omega^{v_1}_{\boldsymbol{x}}|\Omega^{v_2}_{\boldsymbol{x}})$} indicates that the extracted salient concepts have higher transferability across different DNNs.

\begin{figure}[t]
    \centering
    \includegraphics[width=\linewidth]{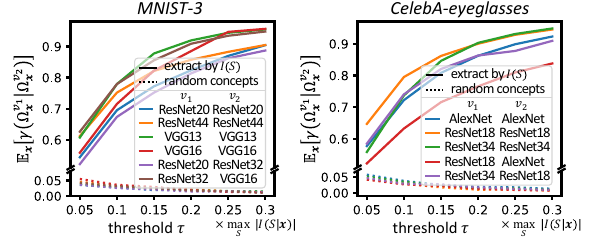}
    \vspace{-23pt}
    \caption{Concepts extracted by a higher threshold {\small$\tau$} (\emph{i.e.} concepts with more significant effects {\scriptsize$I(S|\boldsymbol{x})$}) usually have higher transferability across different DNNs.}
    \label{fig:model-wise-transfer-diff-tau}
    \vspace{-15pt}
\end{figure}

\textbf{Experiments.}
In this experiment, we examined the transferability of concepts in both the case when DNNs {\small$v_1$} and {\small$v_2$} had the same network architecture but were trained with different parameter initializations, and the case when {\small$v_1$} and {\small$v_2$} had different network architectures.
Then, given each sample {\small$\boldsymbol{x}$}, {\small$\Omega^{v_2}_{\boldsymbol{x}}$} contains all salient concepts with interaction strength {\small$I_{v_2}(S|\boldsymbol{x})\geq 0.05\cdot \max_{S}|I_{v_2}(S|\boldsymbol{x})|$}, as defined in Section~\ref{subsubsec:sparsity-single-sample}. 
Whereas, we used different thresholds {\small$\tau$} ranging from {\small$\tau=0.05\cdot \max_{S}|I_{v_1}(S|\boldsymbol{x})|$} to {\small$\tau=0.3\cdot \max_{S}|I_{v_1}(S|\boldsymbol{x})|$} to generate different sets {\small$\Omega^{v_1}_{\boldsymbol{x}}$}.
A larger {\small$\tau$} value usually generated a smaller set of salient concepts with more significant effects.
These concepts were more likely to be stably learned by different DNNs.
Fig.~\ref{fig:model-wise-transfer-diff-tau} shows that concepts with higher saliency usually exhibited higher transferability from DNN {\small$v_1$} to DNN {\small$v_2$}.
This indicated that more salient concepts were more likely to be stably learned by different DNNs, which was aligned with intuition.
As a baseline for comparison, we also randomly extracted two sets of concepts {\small$\tilde{\Omega}^{v_1}_{\boldsymbol{x}}$} and {\small$\tilde{\Omega}^{v_2}_{\boldsymbol{x}}$} from all the {\small$2^n$} interaction concepts, which had the same size as {\small$\Omega^{v_1}_{\boldsymbol{x}}$} and {\small$\Omega^{v_2}_{\boldsymbol{x}}$}, \emph{i.e.} {\small$|\Omega^{v_1}_{\boldsymbol{x}}|=|\tilde{\Omega}^{v_1}_{\boldsymbol{x}}|$} and {\small$|\Omega^{v_2}_{\boldsymbol{x}}|=|\tilde{\Omega}^{v_2}_{\boldsymbol{x}}|$}.
Fig.~\ref{fig:model-wise-transfer-diff-tau} shows that the transferability {\small$\mathbb{E}_{\boldsymbol{x}}[\gamma(\Omega^{v_1}_{\boldsymbol{x}}|\Omega^{v_2}_{\boldsymbol{x}})]$} of concepts extracted by {\small$I(S|\boldsymbol{x})$} increased in the range of 0.5-0.95 along with the increase of {\small$\tau$}.
In comparison, the transferability {\small$\mathbb{E}_{\boldsymbol{x}}[\gamma(\tilde{\Omega}^{v_1}_{\boldsymbol{x}}|\tilde{\Omega}^{v_2}_{\boldsymbol{x}})]$} of random concepts was usually less than 0.05.
This verified the high transferability of concepts across different DNNs.

\subsubsection{Discrimination power of concepts}
\label{subsubsec:concept-discrmination}

Furthermore, if a DNN encodes faithful symbolic concepts, then these concepts are supposed to exhibit certain discrimination power in the classification task.
In other words, \textbf{for each concept {\small$S$}, if the concept is saliently activated on a set of samples, then interaction effects {\small$I(S)$} of the same concept are supposed to push the classification of these samples towards a certain category in most cases.} 
Note that different concepts extracted from a sample may push the sample towards different categories, and the classification is the result of the competition between these concepts.

In order to verify the above discrimination power of concepts, in this section, we extract the concept {\small$S$} from {\small$m$} different input samples {\small$\boldsymbol{x}_1,\boldsymbol{x}_2,...,\boldsymbol{x}_m$} in the same category {\small$c$}, and check whether this concept consistently exhibits a positive (or negative) interaction effects {\small$I(S)$} on the {\small$m$} samples.
If the concept {\small$S$} pushes the classification of most of the {\small$m$} samples towards the target category, \emph{i.e.,} {\small$I(S|\boldsymbol{x}_i)>0$} (or opposite to the target category, \emph{i.e.,} {\small$I(S|\boldsymbol{x}_i)<0$}), then the discrimination power of the concept {\small$S$} is high.
On the other hand, if the concept {\small$S$} pushes half of the samples towards the positive direction {\small$I(S|\boldsymbol{x}_i)>0$}, but pushes the other half towards the negative direction {\small$I(S|\boldsymbol{x}_i)<0$}, then the discrimination power of the concept {\small$S$} is low.

Specifically, we use the following metric to measure the discrimination power of concept {\small$S$} among the above {\small$m$} samples in category {\small$c$}.
Let {\small$\Omega_{\boldsymbol{x}_i}$} denote a set of salient concepts extracted from the sample {\small$\boldsymbol{x}_i$}.
Then, {\small$m^+_S\triangleq \sum_{i}\mathbbm{1}_{S\in\Omega_{\boldsymbol{x}_i}} \cdot \mathbbm{1}_{I(S|\boldsymbol{x}_i)>0}$} denote the number of samples where the concept {\small$S$} makes a salient and positive effect on the classification score.
 Similarly, we can define {\small$m^-_S\triangleq \sum_{i}\mathbbm{1}_{S\in\Omega_{\boldsymbol{x}_i}} \cdot \mathbbm{1}_{I(S|\boldsymbol{x}_i)<0}$} to denote the number of samples where the concept {\small$S$} makes a salient and negative effect on the classification score.
In this way, the discrimination power of a salient concept {\small$S$} can be measured as {\small$\beta(S)=\max(m^+_S, m^-_S)/(m^+_S + m^-_S)$}.
A larger value of {\small$\beta(S)$} indicates a higher discrimination power of the concept {\small$S$}.

\textbf{Experiments.}
Note that different concepts are of different importance in the classification of a category.
Some concepts frequently appear in different samples and make salient effects, while other concepts only appear in very few concepts. 
Therefore, we use the frequency of the concept {\small$\alpha(S)$} as a weight to compute the average discrimination power of all concepts, which is given as {\small$\bar{\beta}\triangleq\sum_{S}[\alpha(S)\cdot\beta(S)]/\sum_{S}[\alpha(S)]$}.
The frequency of a concept is defined as {\small$\alpha(S)\triangleq (m^+_S+m^-_S)/m$}.
The selection of datasets and the training of DNNs are introduced in the \textit{experimental settings} paragraph at the end of Section~\ref{sec:emergence}.
Fig.~\ref{fig:concept-discrimination-power} shows the average discrimination power of concepts in different frequency intervals.
We found that the average discrimination power {\small$\bar{\beta}$} of concepts was usually higher than {\small$0.8$}, which verified the discrimination power of extracted concepts.

\subsection{When DNNs do not learn transferable concepts}

It is worth noting that all the above work just conducts experiments to show the concept-emerging phenomenon in different DNNs for different tasks.
We do not provide, or there may even do not exist, a theoretical proof for such a concept-emerging phenomenon, although the concept-emerging phenomenon \textbf{does exist} in DNNs for most applications.
Therefore, in this subsection, we would like to discuss the following three extreme cases, in which a DNN does not learn symbolic concepts.

In the three extreme cases, DNNs may either collapse to simple models close to linear regressions, or learn non-transferable indiscriminative concepts, although the network output can still be represented as the sum of interaction effects of these concepts.

$\bullet$~\textbf{Case 1: When there exists label noise.}
If the ground-truth label for classification is incorrectly annotated on some samples, then the DNN usually has to memorize each incorrectly-labeled training sample for classification without summarizing many common features from such chaotic annotations.
Thus, in this case, the DNN usually encodes more non-transferable concepts.

\begin{figure}[t]
    \centering
    \includegraphics[width=.95\linewidth]{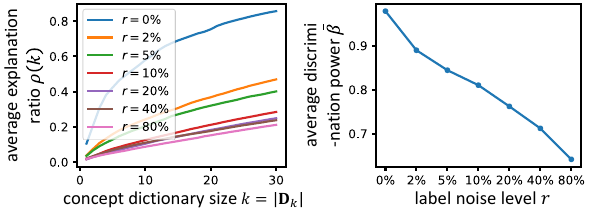}
    \vspace{-10pt}
    \caption{The (left) transferability and (right) discrimination power of concepts decreased when we added more label noises.}
    \label{fig:case-1-label-noise}
    \vspace{-15pt}
\end{figure}

\textit{Experimental verification.}
In this experiment, we constructed datasets with noisy labels to check whether DNNs trained on these datasets did not learn transferable concepts with high discrimination power.
To this end, given a clean dataset, we first selected and randomly labeled a certain portion {\small$r$} of training samples in the dataset, so as to train a DNN.
Specifically, we constructed a series of datasets by assigning different ratios {\small$r$} of samples with incorrect labels.
We used the \textit{wifi} dataset\footref{fn:exp-setting} to construct new datasets by adding different ratios {\small$r$} of noisy labels.
Then, we trained an MLP-5 network\footref{fn:exp-setting} on each of these datasets.
We examined the transferability and discrimination power of the extracted concepts\footref{fn:specific-category} on each MLP-5 network.
We extracted concept dictionaries of different sizes based on each MLP-5 network (please see Section~\ref{subsubsec:trans-different-sample} for details).
Fig.~\ref{fig:case-1-label-noise}(left) shows that if there was significant label noise in the dataset, the concept dictionary usually explained much fewer concepts encoded by the network, which indicated low transferability of the learned concepts.
Besides, Fig.~\ref{fig:case-1-label-noise}(right) shows the average discrimination power {\small$\bar{\beta}$} of the extracted concepts also decreased when we assigned more training samples with random labels.
This verified that the DNN usually could not learn transferable and discriminative concepts from samples that were incorrectly labeled.

$\bullet$~\textbf{Case 2: When input samples are noisy.}
In fact, this case can be extended to a more general scenario, \emph{i.e.}, when the task is difficult to learn, there is no essential difference between the difficult data and noisy data for the DNN.
Specifically, when training samples are noisy and lack meaningful patterns, it is difficult for a DNN to learn transferable concepts from noisy training samples.

\textit{Experimental verification.}
In this experiment, we injected noise into training samples to examine whether DNNs trained on datasets with noisy samples did not learn transferable concepts.
Just like the experiment in ``Case 1,'' we constructed such datasets by corrupting a clean dataset.
To this end, we added Gaussian noises {\small$\boldsymbol{\epsilon}\sim\mathcal{N}(\mathbf{0}, \mathbf{I})$} to each input sample {\small$\boldsymbol{x}$} in the clean dataset by modifying it to {\small$(1-\delta)\cdot\boldsymbol{x}+\delta\cdot\boldsymbol{\epsilon}$}, where {\small$\delta\in[0,1]$} denotes the strength of noise injected into the sample {\small$\boldsymbol{x}$}.
Each dimension of the clean sample {\small$\boldsymbol{x}$} was normalized to unit variance over the dataset beforehand.

\begin{figure}[t]
    \centering
    \includegraphics[width=.95\linewidth]{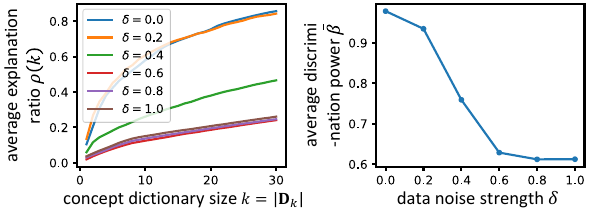}
    \vspace{-10pt}
    \caption{The (left) transferability and (right) discrimination power of concepts decreased when input data were noisy.}
    \label{fig:case-2-data-noise}
    \vspace{-15pt}
\end{figure}

We constructed a series of datasets by injecting noises of different strength {\small$\delta$} into samples in the \textit{wifi} dataset\footref{fn:exp-setting}.
We trained MLP-5's\footref{fn:exp-setting} based on these datasets.
Just like experiments in ``Case 1,'' we examined the transferability and discrimination power of concepts.
Fig.~\ref{fig:case-2-data-noise}(left) shows that the transferability of concepts was usually low when the DNN was learned from noisy input data.
Fig.~\ref{fig:case-2-data-noise}(right) shows that the average discrimination power {\small$\bar{\beta}$} of concepts decreased along with the increasing strength {\small$\delta$} of injected noise.
This verified that DNN usually did not learn transferable and discriminative concepts when input data were noisy.

$\bullet$~\textbf{Case 3: When the task has a simple shortcut solution.}
In the above two cases, both label noise and data noise corrupted the original discriminative patterns in each category, thus making the DNN unlikely to learn transferable concepts.
In comparison, here, let us discuss a new case, \emph{i.e.}, even if there exist meaningful patterns in training data, the DNN may still not learn these concepts.

To be precise, if a classification task can be conducted with some shortcut solutions without requiring the DNN to encode complex concepts, then the DNN probably converges to the shortcut solution.
For example, in an image classification task, if pixel-wise colors are sufficient to conduct the image-classification task, then the DNN is more likely to only use the color information for classification without modeling complex visual concepts.
The simple shortcut solution usually prevents the DNN from summarizing complex visual concepts.

\begin{figure}[t]
\begin{minipage}{.42\linewidth}
    \centering
    \includegraphics[width=.98\linewidth]{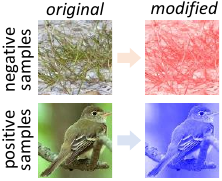}
\end{minipage}
\hfill
\renewcommand{\arraystretch}{1.2}
\begin{minipage}{.57\linewidth}
\centering
\resizebox{\linewidth}{!}{%
\begin{tabular}{c|cc}
\hline
\multirow{4}{*}{\vspace{4pt}\begin{tabular}[c]{@{}c@{}}Relative\\[-3pt] concept\\[-3pt] strength {\small$\kappa$}\end{tabular}} & \multicolumn{2}{c}{Dataset} \\ \cline{2-3} 
 & the original & the modified \\[-3pt]
 & \textit{CUB-binary} & \textit{CUB-binary} \\[-3pt] 
 & dataset & dataset \\ \hline
AlexNet & 0.32 & 0.05 \\ \hline
ResNet-18 & 0.20 & 0.07 \\ \hline
VGG-13 & 0.24 & 0.04 \\ \hline
\end{tabular}%
}
\end{minipage}
\renewcommand{\arraystretch}{1.0}
\vspace{-12pt}
\caption{(left) We constructed a dataset where the ``\textit{color}'' information was a shortcut solution. (right) The relative concept strength {\small$\kappa$} extracted from DNNs trained on different datasets.}
\label{fig:shortcut-case}
\vspace{-10pt}
\end{figure}

\textit{Experimental verification.}
We constructed a dataset for image classification, where the \textit{``color''} information was a shortcut solution for the task.
Specifically, we modified images in the \textit{CUB-binary} dataset\footref{fn:exp-setting}, such that all negative samples were red-colored background regions, and all positive samples were blue-colored foreground birds, as shown in Fig.~\ref{fig:shortcut-case}(left).
We trained AlexNet, ResNet-18, and VGG-13 on both the original dataset and the modified dataset.
Compared with DNNs learned on the original dataset, DNNs learned on the modified dataset were more likely to simply used the color information for classification.
We used the metric {\small$\kappa\triangleq\mathbb{E}_{\boldsymbol{x}}[\sum_{S\in\Omega_{\boldsymbol{x}},|S|\geq 2}|I(S)|/\sum_{S\in\Omega_{\boldsymbol{x}}}|I(S)|]$} to measure the relative strength of all concepts consisting of multiple variables.
Fig.~\ref{fig:shortcut-case}(right) shows that the {\small$\kappa$} values were usually low for DNNs learned to classify red-colored backgrounds and blue-colored birds.
This indicated that the DNN collapsed to a simple model without encoding interactions between different image patches when the task had a simple shortcut solution.

\vspace{-2pt}
\textbf{Experimental settings.}
For tabular datasets, we used the UCI tic-tac-toe endgame dataset~\cite{Dua:2019} for binary classification, and used the UCI wireless indoor localization dataset~\cite{Dua:2019} for multi-category classification.
These datasets were termed \textit{tic-tac-toe} and \textit{wifi} for simplicity.
We trained the following two MLPs on each tabular dataset.
\textit{MLP-5} contained five fully connected layers with 100 neurons in each hidden layer~\cite{ren2021towards}.
\textit{ResMLP-5} was constructed by adding a skip connection to each layer of an \textit{MLP-5}.
For image data, we used the following three datasets.
We took images corresponding to digit ``three'' in the MNIST dataset~\cite{lecun1998mnist} as positive samples, and took other images as negative samples to train DNNs.
We took images with the attribute ``eyeglasses'' in the CelebA dataset~\cite{liu2015faceattributes} as positive samples, and took other images as negative samples to train DNNs.
We trained DNNs to classify birds in bounding boxes in the CUB-200-2011 dataset~\cite{wah2011caltech} from randomly cropped background regions around the bird.
These three datasets were termed \textit{MNIST-3}, \textit{CelebA-eyeglasses}, and \textit{CUB-binary} for short.
We trained LeNet~\cite{lecun1998gradient}, AlexNet~\cite{krizhevsky2017imagenet}, ResNet-18/20/32/34/44~\cite{he2016deep}, VGG-13/16~\cite{simonyan2014very} on these image datasets.
Based on the ShapeNet dataset~\cite{yi2016scalable} for the classification of 3D point clouds, we trained PointNet~\cite{qi2017pointnet} and PointNet++~\cite{qi2017pointnet++}.
Please see Appendix~\ref{app:accuracy} for the classification accuracy of the above DNNs.

\vspace{-2pt}
\section{Conclusion, discussions, and future challenges}

In this paper, we have analyzed the interaction concepts encoded by a DNN.
Specifically, we quantitatively examine the concept-emerging phenomenon of a DNN from four perspectives.
Extensive empirical studies have verified that a well-trained DNN usually encodes sparse, transferable, and discriminative interaction concepts.
Our experiments also prove the faithfulness of the interaction concepts extracted from DNNs. 
Besides, we also discussed three cases in which a DNN is unlikely to learn transferable concepts.

Despite the achievements in this study, it is still hard to say that the interaction concept is an essential explanation of the DNN.
Several big scientific problems are still unsolved in this field.
For example, using interaction concepts to explain a neural network's learning dynamics, identifying memorization effects and reasoning effects used by the DNN, evaluating the correctness of a DNN's detailed decision-making logic, and extracting generalizable interaction concepts are all typical challenges.

\textbf{Acknowledgement.}
This work is partially supported by the National Nature Science Foundation of China (62276165), National Key R\&D Program of China (2021ZD0111602), Shanghai Natural Science Foundation (21JC1403800,21ZR1434600), National Nature Science Foundation of China (U19B2043).
This work is also partially supported by Huawei Technologies Inc.

\bibliography{ref}
\bibliographystyle{icml2023}

\newpage
\appendix
\onecolumn

\section{Axioms and theorems of the Harsanyi dividend}
\label{app:harsanyi-axiom-theorem}

As mentioned in Section~\ref{subsec:preliminaries} of the paper, the definition for an interaction concept {\small$S$} in Eq. (\ref{eq:harsanyi-def}) can be understood as the Harsanyi dividend of the set of variables in {\small$S$} \emph{w.r.t.} the DNN {\small$v$}.
In fact, the Harsanyi dividend {\small$I(S|\boldsymbol{x})$} also satisfies many desirable axioms and theorems, as follows.

The Harsanyi dividend {\small$I(S|\boldsymbol{x})$} satisfies seven desirable axioms in game theory~\cite{ren2021towards}, including the \textit{efficiency, linearity, dummy, symmetry, anonymity, recursive} and \textit{interaction distribution} axioms.

(1) \textit{Efficiency axiom}. The output score of a model can be decomposed into interaction effects of different patterns, \emph{i.e.} {\small $v(\boldsymbol{x})=\sum_{S\subseteq N}I(S|\boldsymbol{x})$}.

(2) \textit{Linearity axiom}. If we merge output scores of two models $w$ and $v$ as the output of model $u$, \emph{i.e.} {\small $\forall S\subseteq N,~ u(\boldsymbol{x}_S)=w(\boldsymbol{x}_S)+v(\boldsymbol{x}_S)$}, then their interaction effects {\small $I_v(S|\boldsymbol{x})$} and {\small $I_w(S|\boldsymbol{x})$} can also be merged as {\small $\forall S\subseteq N, I_u(S|\boldsymbol{x})=I_v(S|\boldsymbol{x})+I_w(S|\boldsymbol{x})$}.

(3) \textit{Dummy axiom}. If a variable {\small $i\in N$} is a dummy variable, \emph{i.e.} 
{\small $\forall S\subseteq N\backslash\{i\}, v(\boldsymbol{x}_{S\cup\{i\}})=v(\boldsymbol{x}_S)+v(\boldsymbol{x}_{\{i\}})$}, then it has no interaction with other variables, {\small$\forall \emptyset\neq T\subseteq N\backslash\{i\}$}, {\small$I(T\cup\{i\}|\boldsymbol{x})=0$}.

(4) \textit{Symmetry axiom}. If input variables {\small $i,j\in N$} cooperate with other variables in the same way, {\small $\forall S\subseteq N\backslash\{i,j\}, v(\boldsymbol{x}_{S\cup\{i\}})=v(\boldsymbol{x}_{S\cup\{j\}})$}, then they have same interaction effects with other variables, {\small $\forall S\subseteq N\backslash\{i,j\}, I(S\cup\{i\}|\boldsymbol{x})=I(S\cup\{j\}|\boldsymbol{x})$}.

(5) \textit{Anonymity axiom}. For any permutations $\pi$ on {\small $N$}, we have {\small $\forall S
\!\subseteq\! N, I_{v}(S|\boldsymbol{x})=I_{\pi v}(\pi S|\boldsymbol{x})$}, where {\small $\pi S \!\triangleq\!\{\pi(i)|i\!\in\! S\}$}, and the new model {\small $\pi v$} is defined by {\small $(\pi v)(\boldsymbol{x}_{\pi S})=v(\boldsymbol{x}_{S})$}.
This indicates that interaction effects are not changed by permutation.

(6) \textit{Recursive axiom}. The interaction effects can be computed recursively.
For {\small $i\in N$} and {\small $S\subseteq N\backslash\{i\}$}, the interaction effect of the pattern {\small $S\cup\{i\}$} is equal to the interaction effect of {\small $S$} with the presence of $i$ minus the interaction effect of $S$ with the absence of $i$, \emph{i.e.} {\small $\forall S\!\subseteq\! N\!\setminus\!\{i\}, I(S\cup \{i\}|\boldsymbol{x})\!=\!I(S|i\text{ is always present},\boldsymbol{x})\!-\!I(S|\boldsymbol{x})$}. {\small $I(S|i\text{ is always present},\boldsymbol{x})$} denotes the interaction effect when the variable $i$ is always present as a constant context, \emph{i.e.} {\small $
I(S|i\text{ is always present},\boldsymbol{x})=\sum_{L\subseteq S} (-1)^{|S|-|L|}\cdot v(\boldsymbol{x}_{L\cup\{i\}})$}.

(7) \textit{Interaction distribution axiom}. This axiom characterizes how interactions are distributed for ``interaction functions''~\cite{sundararajan2020shapley}.
An interaction function {\small $v_T$} parameterized by a subset of variables {\small $T$} is defined as follows.
{\small $\forall S\subseteq N$}, if {\small $T\subseteq S$}, {\small$v_T(\boldsymbol{x}_S)=c$}; otherwise, {\small $v_T(\boldsymbol{x}_S)=0$}.
The function {\small$v_T$} models pure interaction among the variables in {\small$T$}, because only if all variables in {\small$T$} are present, the output value will be increased by {\small$c$}.
The interactions encoded in the function {\small$v_T$} satisfies {\small $I(T|\boldsymbol{x})=c$}, and {\small $\forall S\neq T$}, {\small $I(S|\boldsymbol{x})=0$}.

The Harsanyi dividend {\small$I(S|\boldsymbol{x})$} can also explain the elementary mechanism of existing game-theoretic metrics~\cite{ren2021towards}, including \textit{the Shapley value}, \textit{the Shapley interaction index}, and \textit{the Shapley-Taylor interaction index}.

(1) \textit{Connection to the Shapley value~\cite{shapley1953value}}.
Let {\small $\phi(i|\boldsymbol{x})$} denote the Shapley value of an input variable {\small$i$}, given the input sample {\small$\boldsymbol{x}$}.
Then, the Shapley value {\small$\phi(i|\boldsymbol{x})$} can be explained as the result of uniformly assigning attributions of each Harsanyi dividend to each involving variable {\small$i$}, \emph{i.e.}, {\small $\phi(i|\boldsymbol{x})=\sum_{S\subseteq N\backslash\{i\}}\frac{1}{|S|+1} I(S\cup\{i\}|\boldsymbol{x})$}.
This also proves that the Shapley value is a fair assignment of attributions from the perspective of Harsanyi dividend.

(2) \textit{Connection to the Shapley interaction index~\cite{grabisch1999axiomatic}}.
Given a subset of variables {\small $T\subseteq N$} in an input sample {\small$\boldsymbol{x}$}, the Shapley interaction index {\small$I^{\textrm{Shapley}}(T|\boldsymbol{x})$} can be represented as {\small $I^{\textrm{Shapley}}(T|\boldsymbol{x})=\sum_{S\subseteq N\backslash T}\frac{1}{|S|+1}I(S\cup T|\boldsymbol{x})$}.
In other words, the index {\small$I^{\textrm{Shapley}}(T|\boldsymbol{x})$} can be explained as uniformly allocating {\small$I(S'|\boldsymbol{x})$} s.t. {\small$S'=S\cup T$} to the compositional variables of {\small$S'$}, if we treat the coalition of variables in {\small $T$} as a single variable.

(3) \textit{Connection to the Shapley Taylor interaction index~\cite{sundararajan2020shapley}}.
Given a subset of variables {\small $T\subseteq N$} in an input sample {\small$\boldsymbol{x}$}, the {\small$k$}-th order Shapley Taylor interaction index {\small $I^{\textrm{Shapley-Taylor}}(T|\boldsymbol{x})$} can be represented as weighted sum of interaction effects, \emph{i.e.}, {\small $I^{\textrm{Shapley-Taylor}}(T|\boldsymbol{x})=I(T|\boldsymbol{x})$} if {\small $|T|<k$}; {\small $I^{\textrm{Shapley-Taylor}}(T|\boldsymbol{x})=\sum_{S\subseteq N\backslash T}\binom{|S|+k}{k}^{-1}I(S\cup T|\boldsymbol{x})$} if {\small $|T|=k$}; and {\small $I^{\textrm{Shapley-Taylor}}(T|\boldsymbol{x})=0$ if $|T|>k$}.

\section{Experimental details}

\subsection{Accuracy of DNNs}
\label{app:accuracy}

In this paper, we conducted experiments on various DNNs trained on different types of datasets, including tabular datasets, image datasets, and a point-cloud dataset.
Table \ref{tab:accuracy} reports the classification accuracy of DNNs trained on the above datasets.

\begin{table}[htbp]
\centering
\caption{Classification accuracy of different DNNs.}
\label{tab:accuracy}
\vspace{3pt}
\resizebox{.65\textwidth}{!}{%
\begin{tabular}{c|cccccc}
\hline
Dataset & \multicolumn{6}{c}{DNN} \\ \hline
\multirow{2}{*}{\textit{tic-tac-toe}} & MLP-5 & ResMLP-5 &  &  &  &  \\
 & 100\% & 100\% &  &  &  &  \\ \hline
\multirow{2}{*}{\textit{wifi}} & MLP-5 & ResMLP-5 &  &  &  &  \\
 & 97.75\% & 97.75\% &  &  &  &  \\ \hline
\multirow{2}{*}{\textit{MNIST-3}} & LeNet & ResNet-20 & ResNet-32 & ResNet-44 & VGG-13 & VGG-16 \\
 & 99.99\% & 100\% & 100\% & 100\% & 100\% & 100\% \\ \hline
\multirow{2}{*}{\textit{CelebA-eyeglasses}} & AlexNet & ResNet-18 & VGG-13 &  &  &  \\
 & 99.53\% & 99.66\% & 99.65\% &  &  &  \\ \hline
\multirow{2}{*}{\textit{CUB-binary}} & AlexNet & ResNet-18 & ResNet-34 &  &  &  \\
 & 95.67\% & 96.41\% & 96.43\% &  &  &  \\ \hline
\multirow{2}{*}{\begin{tabular}[c]{@{}c@{}}the modified\\ \textit{CUB-binary} dataset\end{tabular}} & AlexNet & ResNet-18 & ResNet-34 &  &  &  \\
 & 100\% & 100\% & 100\% &  &  &  \\ \hline
\multirow{2}{*}{\textit{ShapeNet}} & PointNet & PointNet++ &  &  &  &  \\
 & 97.36\% & 98.64\% &  &  &  &  \\ \hline
\end{tabular}%
}
\end{table}

\subsection{The annotation of semantic parts}
\label{app:annotation}

This section discusses the annotation of semantic parts in the point-cloud dataset and image datasets.
As mentioned in Section~\ref{subsubsec:sparsity-single-sample}, given an input sample {\small$\boldsymbol{x}$} with {\small$n$} input variables, the DNN may encode at most {\small$2^n$} interaction concepts.
The computational cost for extracting salient concepts is high, when the number of input variables {\small$n$} is large.
For example, if we take each 3D point of a point-cloud (or each pixel of an image) as a single input variable, the computation is usually prohibitive.
In order to overcome this issue, we simply annotate 8-10 semantic parts in each input sample, such that the annotated semantic parts are aligned over different samples\footnote{Actually, we can extract sparse and transferable interaction concepts without pre-annotated parts. Please see Appendix~\ref{app:wo-annotation} for experimental results.}.
Then, each semantic part in an input sample is taken as a ``single'' input variable to the DNN.

\begin{figure}[htbp]
    \centering
    \includegraphics[width=.95\linewidth]{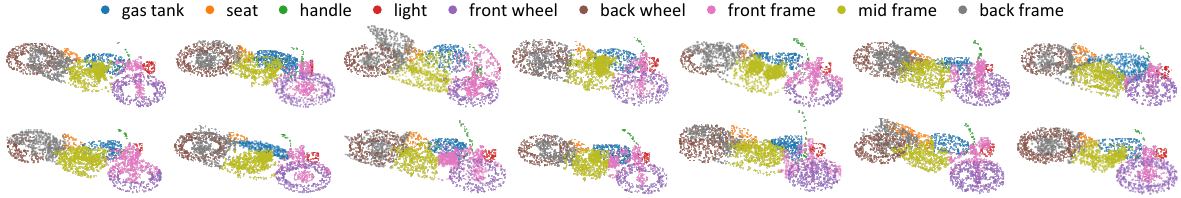}
    \vspace{-5pt}
    \caption{Examples of annotated semantic parts for samples in the \textit{motorbike} category of the ShapeNet dataset.}
    \label{fig:annotation-shapenet}
\end{figure}

$\bullet$~For point-cloud data in the ShapeNet dataset, we annotated semantic parts for 100 samples in the \textit{motorbike} category.
These semantic parts were generated based on original annotations provided by~\cite{yi2016scalable}.
In the original annotation, \citet{yi2016scalable} provided semantic parts including \textit{gas tank}, \textit{seat}, \textit{handle}, \textit{light}, \textit{wheel}, and \textit{frame} for each \textit{motorbike} sample.
As shown in Fig.~\ref{fig:annotation-shapenet}, we further modified the original annotation into more fine-grained semantic parts, \emph{i.e.} \textit{gas tank}, \textit{seat}, \textit{handle}, \textit{light}, \textit{front wheel}, \textit{back wheel}, \textit{front frame}, \textit{mid frame}, and \textit{back frame}.

\begin{figure}[htbp]
    \centering
    \includegraphics[width=.95\linewidth]{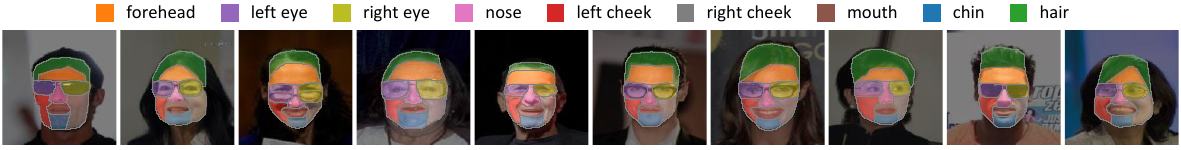}
    \vspace{-5pt}
    \caption{Examples of annotated semantic parts for positive samples in the \textit{CelebA-eyeglasses} dataset.}
    \label{fig:annotation-celeba}
\end{figure}

$\bullet$~For image data, we annotated semantic parts for 50 samples in the \textit{CelebA-eyeglasses} dataset.
Specifically, as shown in Fig.~\ref{fig:annotation-celeba}, we annotated semantic parts including \textit{forehead}, \textit{left eye}, \textit{right eye}, \textit{nose}, \textit{left cheek}, \textit{right cheek}, \textit{mouth}, \textit{chin}, and \textit{hair} for each sample in the \textit{CelebA-eyeglasses} dataset\footnote{Note that we only considered interactions within foreground regions in each image, due to the high computational cost mentioned above.
Therefore, the annotated semantic parts did not cover regions in the background.
Please see Section~\ref{app:image-bg} for details on how to handle background regions in the computation of {\scriptsize$I(S|\boldsymbol{x})$} for image data.\label{fn:image-bg}}.
Similarly, we annotated semantic parts for 20 samples in the \textit{CUB-binary} dataset. These semantic parts include \textit{head}, \textit{neck}, \textit{throat}, \textit{wing}, \textit{tail}, \textit{leg}, \textit{belly}, and \textit{breast}. For images in the \textit{MNIST-3} dataset, we annotated semantic parts for 100 positive samples. Please see the source code for details.

\subsection{The setting of {\small$v(\boldsymbol{x})$} in experiments}
\label{app:v-setting}

As mentioned in Section~\ref{subsec:preliminaries} of the paper, in the computation of the interaction effect {\small$I(S|\boldsymbol{x})$}, people can apply different settings for {\small$v(\boldsymbol{x})$}.
For example, \citet{covert2020understanding} computed {\small$v(\boldsymbol{x})$} as the cross-entropy loss of the sample {\small$\boldsymbol{x}$} in the classification task.
\citet{lundberg2017unified} directly set {\small$v(\boldsymbol{x})=p(y=y^{\text{truth}}|\boldsymbol{x})\in\mathbb{R}$}.
In this paper, we followed~\cite{deng2022discovering} and used {\small$v(\boldsymbol{x})=\log\frac{p(y=y^{\text{truth}}|\boldsymbol{x})}{1-p(y=y^{\text{truth}}|\boldsymbol{x})}\in\mathbb{R}$} for both binary classification tasks and multi-category classification tasks.

\subsection{Computation details of {\small$I(S|\boldsymbol{x})$} for image data}
\label{app:image-bg}

As mentioned in Section~\ref{app:annotation}, given an input sample {\small$\boldsymbol{x}$} with {\small$n$} input variables, the DNN may encode at most {\small$2^n$} interaction concepts.
The computational cost for extracting salient concepts is high, when the number of input variables {\small$n$} is large.
In order to overcome this issue, we only considered interaction concepts formed by foreground regions.
For image data, the annotated semantic parts for each sample only covered regions in the foreground.
In order to handle the uncovered regions in the background, in the extraction of interaction concepts, we averaged the interaction effect {\small$I(S|\boldsymbol{x})$} when we were given multiple background with different strengths, which was similar to~\cite{sundararajan2017axiomatic}.

Specifically, let the input image {\small$\boldsymbol{x}\in\mathbb{R}^{n}$} be divided into the foreground region {\small$\boldsymbol{x}^{\text{fg}}\in\mathbb{R}^{n^{\text{fg}}}$} and the background region {\small$\boldsymbol{x}^{\text{bg}}\in\mathbb{R}^{n^{\text{bg}}}$}, where {\small$n=n^{\text{fg}}+n^{\text{bg}}$} and {\small$\boldsymbol{x}=\boldsymbol{x}^{\text{fg}}\sqcup\boldsymbol{x}^{\text{bg}}$}.
The foreground region {\small$\boldsymbol{x}^{\text{fg}}$} consisted of all pixels covered by the annotated semantic parts in Section~\ref{app:annotation}, and the background region {\small$\boldsymbol{x}^{\text{bg}}$} consisted of all other uncovered pixels.
Let {\small$\boldsymbol{b}^{\text{bg}}\in\mathbb{R}^{n^{\text{bg}}}$} denote the baseline value for pixels in the background region {\small$\boldsymbol{x}^{\text{bg}}\in\mathbb{R}^{n^{\text{bg}}}$}.
We defined the background region {\small$\boldsymbol{x}^{\text{bg}}_{\alpha}$} with strength {\small$\alpha$} \emph{w.r.t.} the baseline value {\small$\boldsymbol{b}^{\text{bg}}$} as {\small$\boldsymbol{x}^{\text{bg}}_{\alpha}=\alpha\cdot\boldsymbol{x}^{\text{bg}}+(1-\alpha)\cdot\boldsymbol{b}^{\text{bg}}$}, where the strength {\small$\alpha\in[0,1]$}.
When {\small$\alpha=0$}, the background region was masked by the baseline value, \emph{i.e.} {\small$\boldsymbol{x}^{\text{bg}}_{\alpha=0}=\boldsymbol{b}^{\text{bg}}$}.
When {\small$\alpha=1$}, the background region remained its original value, \emph{i.e.} {\small$\boldsymbol{x}^{\text{bg}}_{\alpha=1}=\boldsymbol{x}^{\text{bg}}$}.
When we computed the effect of each interaction concept {\small$S$}, we averaged the interaction effect when we were given multiple background regions with different strengths {\small$\alpha$}, as follows.

\vspace{-3pt}
\begin{small}
\begin{equation}
    I(S|\boldsymbol{x})=\mathbb{E}_{\alpha\sim\mathcal{U}[0,1]}\left[I\left(S\ \big|\ \boldsymbol{x}^{\text{fg}}\sqcup\boldsymbol{x}^{\text{bg}}_{\alpha}\right)\right]
    =\int_0^1 I\left(S\ \big|\ \boldsymbol{x}^{\text{fg}}\sqcup\boldsymbol{x}^{\text{bg}}_{\alpha}\right)\ \mathrm{d}\alpha
\end{equation}
\end{small}

\subsection{The eight sub-categories in the \textit{tic-tac-toe} dataset}
\label{app:tictactoe-subcat}

In this section, we provide more details on the eight sub-categories for positive samples in the \textit{tic-tac-toe} dataset, as mentioned in the footnote\footref{fn:specific-category} of the paper.

Each sample {\small$\boldsymbol{x}$} in the \textit{tic-tac-toe} dataset~\cite{Dua:2019} encodes one possible board configurations at the end of tic-tac-toe games.
Specifically, each variable {\small$\boldsymbol{x}_i$} indicates the state at the {\small$i$}-th position of the board, where {\small$\boldsymbol{x}_i=1$} indicates the player ``X'' has taken this position, {\small$\boldsymbol{x}_i=-1$} indicates the player ``O'' has taken this position, and {\small$\boldsymbol{x}_i=0$} indicates this position is blank.
If one of the player in the tic-tac-toe game creates a ``three-in-a-row'', then this player wins the game.
In the \textit{tic-tac-toe} dataset, positive samples includes all configurations where the player ``X'' wins the game.
Since there are eight possible ways for the player ``X'' to create a ``three-in-a-row'', there are eight corresponding sub-categories for positive samples in the \textit{tic-tac-toe} dataset.
Specifically, these sub-categories contain patterns {\small$\boldsymbol{x}_1=\boldsymbol{x}_2=\boldsymbol{x}_3=1$} (three-in-the-first-row),
{\small$\boldsymbol{x}_4=\boldsymbol{x}_5=\boldsymbol{x}_6=1$} (three-in-the-second-row), 
{\small$\boldsymbol{x}_7=\boldsymbol{x}_8=\boldsymbol{x}_9=1$} (three-in-the-third-row), 
{\small$\boldsymbol{x}_1=\boldsymbol{x}_4=\boldsymbol{x}_7=1$} (three-in-the-first-column), 
{\small$\boldsymbol{x}_2=\boldsymbol{x}_5=\boldsymbol{x}_8=1$} (three-in-the-second-column), 
{\small$\boldsymbol{x}_3=\boldsymbol{x}_6=\boldsymbol{x}_9=1$} (three-in-the-third-column), 
{\small$\boldsymbol{x}_1=\boldsymbol{x}_5=\boldsymbol{x}_9=1$} (three-in-the-main-diagonal), 
{\small$\boldsymbol{x}_3=\boldsymbol{x}_5=\boldsymbol{x}_7=1$} (three-in-the-anti-diagonal), respectively.

\section{More experimental results}

\subsection{More verification on the existence of a concept dictionary}
\label{app:concept-dict-more}

\begin{figure*}[htbp]
    \centering
    \includegraphics[width=.95\linewidth]{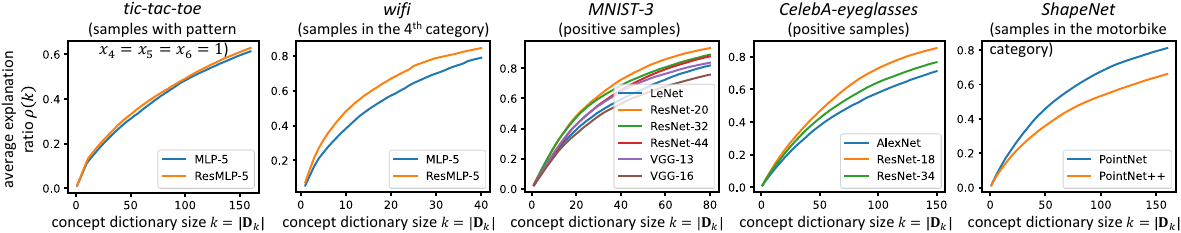}
    \vspace{-5pt}
    \caption{The change of the average explanation ratio {\small$\rho(k)$} along with the size {\small$k$} of the concept dictionary {\small$\mathbf{D}_k$}, when we extracted salient concepts using the vanilla threshold {\scriptsize$\tau=0.05\cdot\max_{S}|I(S|\boldsymbol{x})|$}.}
    \label{fig:verify-concept-dictionary_tau=0.05}
    \vspace{-5pt}
\end{figure*}

As a supplement to Fig.~\ref{fig:verify-concept-dictionary}, Section~\ref{subsubsec:trans-different-sample} of the paper, we conducted another experiment to show the existence of a small concept dictionary {\small$\mathbf{D}_k$} that could explain most concepts encoded by the DNN.
Different from the experiment in Section~\ref{subsubsec:trans-different-sample}, we extracted salient concepts {\small$\Omega_{\boldsymbol{x}}$} by using the vanilla threshold {\small$\tau\!=\!0.05\cdot \max_{S}|I(S|\boldsymbol{x})|$}.
Fig.~\ref{fig:verify-concept-dictionary_tau=0.05} shows that there usually existed a concept dictionary consisting of 40-150 concepts, which could explain more than 60\%-80\% salient concepts encoded by the DNN.
This still verified that the DNN learned transferable concepts over different samples.

\subsection{Extracting interaction concepts without the annotation of semantic parts}
\label{app:wo-annotation}

\begin{figure}[htbp]
    \centering
    \includegraphics[width=.65\linewidth]{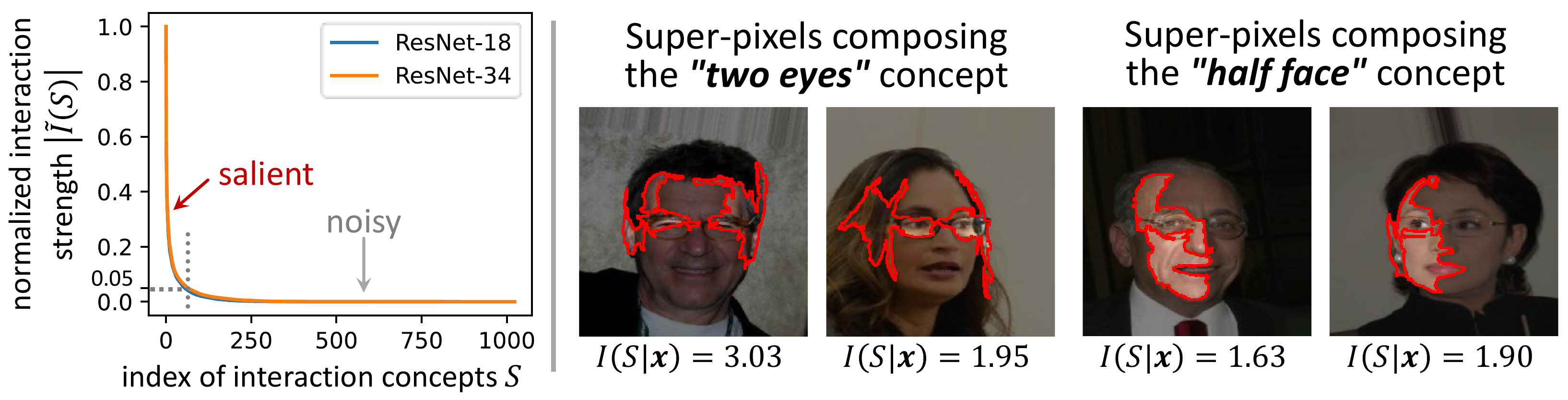}
    \vspace{-5pt}
    \caption{(left) Normalized strength of interaction effects of different concepts in a descending order. Concepts extracted from super-pixels are still sparse. (right) Visualization of salient  concepts  extracted from  ResNet-18. The concepts are usually meaningful to humans, and they are also transferable across different samples.}
    \label{fig:superpixel}
\end{figure}

In this section, we conducted an experiment to extract concepts without pre-defined semantic parts.
In this experiment, we first extracted super-pixels from the image, and consider each super-pixel as a basic input unit of the DNN. 
Then, we can extract interaction concepts between these super-pixels, based on {\small$I(S|x)$} in Eq. (\ref{eq:harsanyi-def}). 
Specifically, we first segmented super-pixels from images in the CelebA dataset using the SLIC method~\cite{achanta2012slic}.
Then, we extracted concepts encoded by ResNet-18 and ResNet-34 trained on the CelebA dataset for the classification of the eyeglasses attribute.
Following the experimental settings in Fig.~\ref{fig:sparsity-single-sample} of the paper, we visualized normalized strength of interaction effects of different concepts in a descending order. 
Experimental results in Fig. \ref{fig:superpixel} show that the extracted concepts were still sparse.
We also visualized some salient concepts extracted from ResNet-18 formed by super-pixels.
We found that the salient concepts were usually meaningful to humans (super-pixels forming the "half face" concept, the "two eyes" concept, \emph{etc}.), and they were also transferable across different samples.

\subsection{Extracting interaction concepts encoded by NLP models}
\label{app:nlp}

In this section, we extracted interaction concepts encoded by DNNs trained on NLP tasks.

In the first experiment, we trained a CNN network~\cite{kim-2014-convolutional} on the SST-2 dataset for the sentiment classification task. 
Then, we extracted interaction concepts from this DNN. Table \ref{tab:nlp-classification} shows the effects of salient concepts to the DNN's output for positive sentiment. 
We found that the extracted concepts were meaningful to human. For example, given the input sentence \textit{"It's just not very smart,"} two salient interaction concepts {\small$\{\textit{not, smart}\}$} and {\small$\{\textit{just, not}\}$} contributed negative scores to the positive sentiment, while the interaction concepts {\small$\{\textit{just, very}\}$} and {\small$\{\textit{just, very, smart}\}$} contributed positive scores to the positive sentiment.

\begin{table}[htbp]
    \centering
    \caption{Effects of salient concepts extracted from a CNN network for the semantic classification task.}
    \label{tab:nlp-classification}
    \vspace{5pt}
    \resizebox{.6\textwidth}{!}{
    \begin{tabular}{c|c}
    \hline
        \begin{tabular}[c]{@{}c@{}}Sentence 1: It's just not very smart.\\ Output: negative sentiment\end{tabular} & \begin{tabular}[c]{@{}c@{}}Sentence 2: It is not too fast and not too slow.\\ Output: positive sentiment\end{tabular} \\ \hline
        $I(\{\text{not, very}\}|x)=-2.56$                    & $I(\{\text{not, too}\}|x)=6.49$ (the first "not too") \\
        $I(\{\text{just, not}\}|x)=-1.52$ & $I(\{\text{not, too}\}|x)=4.87$ (the second "not too") \\
        $I(\{ \text{just, very}\}|x)=0.95$ & $I(\{\text{too}\}|x)=-3.33$ \\ 
        $I(\{\text{just, very, smart}\}|x)=0.84$ & $I(\{\text{not, slow}\}|x)=1.59$ \\
        $I(\{\text{not, smart}\}|x)=-0.77$ & $I(\{\text{and, not}\}|x)=-1.16$ \\ \hline
    \end{tabular}
    }
\end{table}

In the second experiment, we explained concepts encoded by a large language model (OPT-1.3B~\cite{zhang2022opt}) for the text generation task. Given the first {\small$k$} words in the sentence, we focused on the probability distribution of generating the {\small$(k+1)$}-th word. For example, given a partial sentence {\small$x=$}\textit{"Diabetes is a chronic condition that affects how the body uses and stores,"} we focused on the output logits of the next word \textit{"glucoses,"} \emph{i.e.}, {\small$v(x)=\log\frac{p_{next}}{1-p_{next}}$}, where {\small$p_{next}=p(\text{glucoses}|\text{Diabetes ... stores})$}. 
Thus, we extracted interaction concepts with salient effects on generating the target word. 
In Table \ref{tab:nlp-generation}, we showed that the model encoded meaningful concepts. 
For example, in Sentence 1, the model encoded concepts formed by relevant verbs ({\small$\{\textit{affects, and, stores}\}$}), and concepts formed by both relevant nouns and verbs ({\small$\{\textit{body, stores}\}$}). Both the interaction between {\small$\{\textit{affects, and, stores}\}$} and the interaction between {\small$\{\textit{body, stores}\}$} contributed to the correct generation of the output word \textit{"glucoses."}

\begin{table}[htbp]
    \centering
    \caption{Effects of salient concepts extracted from OPT-1.3B for the text generation task.}
    \label{tab:nlp-generation}
    \vspace{5pt}
    \resizebox{.65\textwidth}{!}{
    \begin{tabular}{c|c}
    \hline
        \begin{tabular}[c]{@{}c@{}}Sentence 1: Diabetes is a chronic condition that affects \\ how the body uses and stores, Output: glucoses\end{tabular} & \begin{tabular}[c]{@{}c@{}}Sentence 2: Physicist Isaac newton was born in 1642\\ in the village of Newton, Output: Abbot\end{tabular} \\ \hline
        $I(\{\text{Diabetes}\}|x) = 3.10$ & $I(\{\text{village}\}|x)=2.07$ \\
        $I(\{\text{body, stores}\}|x)=3.08$ & $I(\{\text{village,Newton}\}|x)=1.05$ \\
        $I(\{\text{affects, and, stores}\}|x)=2.62$ & $I(\{\text{Issac,village,Newton}\}|x)=0.90$ \\
        $I(\{\text{Diabetes,body,stores}\}|x)=-2.01$ & $I(\{\text{was}\}|x)=-0.74$ \\
        $I(\{\text{how,body,stores}\}|x)=1.95$ & $I(\{\text{1642,in,village,Newton}\}|x)=-0.71$ \\ \hline
    \end{tabular}
    }
\end{table}

\subsection{Discussion on the relationship between interaction concepts and adversarial robustness}
\label{app:robustness}

In this section, we analyze the relationship between different interaction concepts encoded by the DNN and the adversarial robustness of the DNN.

In the first experiment, we studied the robustness of different concepts.
In this experiment, we found that high-order concepts (\emph{i.e.} concepts which contain massive input variables) were less robust than low-order concepts (\emph{i.e.} concepts which contain a small number of input variables). 
Therefore, we can examine different models based on the extracted concepts, and select models that encode less high-order non-robust concepts. In this way, the selected model would potentially be more robust. 

Mathematically, we defined the \textit{order} of a concept {\small$S$} as the number of input variables composing this concept, \emph{i.e.} {\small$\text{order}(S)=|S|$}. Then, we evaluated the sensitivity of concepts $S$ with different orders {\small$|S|$}, which was encoded by the VGG-16 model trained on the MNIST-3 dataset, when adversarial perturbations~\cite{madry2018towards} were injected into the input sample. Given an input sample {\small$x$}, the adversarial perturbation {\small$\delta$} was obtained via the {\small$L_\infty$} PGD attack~\cite{madry2018towards}, subject to {\small$\Vert\delta\Vert_\infty<\frac{64}{255}$}. The attack was iterated for 20 steps with the step size {\small$\frac{4}{255}$}. The sensitivity of concepts {\small$S$} with {\small$s$}-order (\emph{i.e.} {\small$|S|=s$}) was defined as {\small$\text{sensitivity}_s\triangleq \mathbb{E}_x\left[\frac{\sum_{S:|S|=s}|I(S|x+\delta)-I(S|x)|}{\sum_{S:|S|=s}|I(S|x)|}\right]$}. Table \ref{tab:sensitivity-different-order} shows the sensitivity of concepts with different orders. 

We found that high-order concepts usually exhibited higher sensitivity, thereby being less robust for inference.
Notice that Eq. (\ref{eq:harsanyi-sum}) in the paper shows the network output can be written as the sum of effects of all interaction concepts.
Therefore, if a model encodes massive high-order concepts, the model would probably be less robust to adversarial attacks.
This indicated that we could select models that encode less high-order non-robust concepts. In this way, the selected model would potentially be more robust.

\begin{table}[htbp]
    \centering
    \caption{Sensitivity of concepts with different orders. High-order concepts are sensitive to adversarial noise, thereby being less robust.}
    \label{tab:sensitivity-different-order}
    \vspace{5pt}
    \resizebox{.45\textwidth}{!}{
        \begin{tabular}{c|cccccc}
        \hline
         & {\small$s=1$} & {\small$s=2$} & {\small$s=3$} & {\small$s=4$} & {\small$s=5$} & {\small$s=6$} \\ \hline
        {\small$\text{sensitivity}_s$} & 0.81 & 0.94 & 2.45 & 3.46 & 9.90 & 14.89 \\ \hline
        \end{tabular}
    }
\end{table}

In the second experiment, we compared the transferability of concepts encoded by normally trained models with adversarially trained models~\cite{madry2018towards}. 
We found that besides improving the robustness of the model, adversarial training also improved the generalization power of features, \emph{i.e.}, the transferability of the encoded concepts.
Therefore, we can select models that encoded more transferable concepts, which would potentially be more reliable.

To this end, we trained another two VGG-13 networks and another two VGG-16 networks on the MNIST-3 dataset using adversarial training. Then, following the experimental settings in Fig. \ref{fig:model-wise-transfer-diff-tau}, Section~\ref{subsubsec:trans-different-dnn}, we checked the model-wise transferability of the concepts encoded by these DNNs. Specifically, let us suppose a pair of models {\small$v_1$} and {\small$v_2$} were trained for the same task. Given an input sample {\small$x$}, let {\small$\Omega^{v_1}_x$} and {\small$\Omega^{v_2}_x$} denote the sets of salient concepts extracted by {\small$v_1$} and {\small$v_2$} from sample {\small$x$}, respectively. We evaluated the ratio of concepts in {\small$\Omega^{v_1}_{x}$} encoded by {\small$v_1$} that were also encoded by {\small$v_2$} in {\small$\Omega^{v_2}_{x}$}, \emph{i.e.} {\small$\gamma(\Omega^{v_1}_{x}|\Omega^{v_2}_{x})\triangleq |\Omega^{v_1}_{x}\cap \Omega^{v_2}_{x}|/|\Omega^{v_1}_{x}|$}, to measure the transferability of salient concepts in {\small$\Omega^{v_1}_{\boldsymbol{x}}$}.
Following experimental settings in Section~\ref{subsubsec:trans-different-dnn}, given each sample {\small$x$}, {\small$\Omega^{v_2}_{\boldsymbol{x}}$} contained all salient concepts with interaction strength {\small$I_{v_2}(S|x)\geq 0.05\cdot \max_{S}|I_{v_2}(S|x)|$}. 
We used different thresholds {\small$\tau$} ranging from {\small$\tau=0.05\cdot \max_{S}|I_{v_1}(S|x)|$ to $\tau=0.3\cdot \max_{S}|I_{v_1}(S|x)|$} to generate different sets {\small$\Omega^{v_1}_{x}$}. 
A larger {\small$\tau$} value usually generated a smaller set of salient concepts with more significant effects.
We computed the average ratio over different samples {\small$\mathbb{E}_x[\gamma(\Omega^{v_1}_{x}|\Omega^{v_2}_{x})]$} to measure the transferability of concepts between a pair of models.

Table \ref{tab:adv-trans-v13} and Table \ref{tab:adv-trans-v16} show that concepts encoded by adversarially trained models usually exhibit higher model-wise transferability. This may explain the robustness of adversarially trained models, to some extent. 
Therefore, besides improving the robustness of the model, adversarial training also improved the generalization power of features, \emph{i.e.}, the transferability of the encoded concepts.
This indicated that we could select models that encoded more transferable concepts, which would potentially be more reliable.

\begin{table}[htbp]
    \centering
    \caption{\small Transferability of concepts between a pair of VGG-13 networks, when we extract salient concepts {\small$\Omega^{v_1}_{x}$} under different thresholds.}
    \label{tab:adv-trans-v13}
    \vspace{5pt}
    \resizebox{.7\textwidth}{!}{
    \begin{tabular}{l|cccccc}
    \hline
        {\small$\lambda$}, the threshold {\small$\tau=\lambda\cdot \max_{S}|I_{v_1}(S|x)|$} & 0.05 & 0.10 & 0.15 & 0.20 & 0.25 & 0.30 \\ \hline
        a pair of normally trained VGG-13 networks & 0.61 & 0.77 & 0.87 & 0.91 & 0.94 & 0.95 \\
        a pair of adversarially trained VGG-13 networks & \textbf{0.66} & \textbf{0.80} & 0.87 & \textbf{0.93} & \textbf{0.96} & \textbf{0.98} \\ \hline
    \end{tabular}
    }
\end{table}

\begin{table}[htbp]
    \centering
    \caption{\small Transferability of concepts between a pair of VGG-16 networks, when we extract salient concepts {\small$\Omega^{v_1}_{x}$} under different thresholds.}
    \label{tab:adv-trans-v16}
    \vspace{5pt}
    \resizebox{.7\textwidth}{!}{
    \begin{tabular}{l|cccccc}
    \hline
        {\small$\lambda$}, the threshold {\small$\tau=\lambda\cdot \max_{S}|I_{v_1}(S|x)|$} & 0.05 & 0.10 & 0.15 & 0.20 & 0.25 & 0.30 \\ \hline
        a pair of normally trained VGG-16 networks & 0.56 & 0.71 & 0.82 & 0.88 & \textbf{0.95} & 0.96 \\ 
        a pair of adversarially trained VGG-16 networks & \textbf{0.62} & \textbf{0.76} & \textbf{0.85} & \textbf{0.91} & 0.94 & 0.96 \\ \hline
    \end{tabular}
    }
\end{table}

\end{document}